\documentclass[10pt,twocolumn,letterpaper]{article}

\usepackage{cvpr}
\usepackage{times}
\usepackage{tikz}
\usepackage{pgfplots}
\usepgfplotslibrary{groupplots}
\usepackage{epsfig}
\usepackage{graphicx}
\usepackage{amsmath}
\usepackage{amssymb}
\usepackage{authblk}
\usepackage{subcaption}

\usepackage[breaklinks=true,bookmarks=false]{hyperref}

\cvprfinalcopy 

\def\cvprPaperID{1552} 

\newlength\figureheight
\newlength\figurewidth

\newcommand{\fig}[1]{Figure~\ref{fig:#1}}

\newcommand{\tab}[1]{Table~\ref{tab:#1}}
\newcommand{\eq}[1]{(\ref{eq:#1})}

\ifcvprfinal\fi
\begin{document}

\title{Fast ConvNets Using Group-wise Brain Damage}

\author{Vadim Lebedev\textsuperscript{1,2} \hspace{6cm} Victor Lempitsky\textsuperscript{1}\\
\textsuperscript{1}Skolkovo Institute of Science and Technology (Skoltech), Moscow, Russia\\
\textsuperscript{2}Yandex, Moscow, Russia\\
{\tt\small \{vadim.lebedev, lempitsky\}@skoltech.ru}
}

\maketitle


%

\begin{abstract}
We revisit the idea of brain damage, i.e.\ the pruning of the coefficients of a neural network, and suggest how brain damage can be modified and used to speedup convolutional layers in ConvNets. The approach uses the fact that many efficient implementations reduce generalized convolutions to matrix multiplications. The suggested brain damage process prunes the convolutional kernel tensor in a group-wise fashion. After such pruning, convolutions can be reduced to multiplications of thinned dense matrices, which leads to speedup. We investigate different ways to add group-wise prunning to the learning process, and show that several-fold speedups of convolutional layers can be attained using group-sparsity regularizers. Our approach can adjust the shapes of the receptive fields in the convolutional layers, and even prune excessive feature maps from ConvNets, all in data-driven way.
\end{abstract}

\section{Introduction}
In the original \textit{Optimal Brain Damage} work~\cite{LeCun90} of 25 years ago, LeCun et al.\ observed that a carefully designed ``brain-damage'' process can sparsify the coefficients of a multi-layer neural network very significantly while incurring minimal or no loss of the prediction accuracy. Such process resembles the biological learning processes in mammals, in whose brains the number of synapses peak during early childhood and is then reduced substantially in the process of \textit{synaptic pruning}~\cite{Chechik98}. The optimal brain damage algorithm and its variants, however, impose sparsity in an unstructured way. As a result, while a large number of parameters can be pruned, the attained level of sparsity in the network is usually insufficient to achieve substantial computational speedup.

These days, due to the overwhelming success of very big convolutional neural networks (ConvNets) \cite{LeCun89} on a variety of machine learning problems, the task of speeding up ConvNets has become a topic of active research and engineering. \textit{Generalized convolution}, i.e.\ the operation of convolving a 4D \textit{kernel tensor} with the stack of input maps in order to produce the stack of output maps, is at the core of ConvNets and also represents their speed bottleneck. Here, we present a simple approach that modifies the standard generalized convolution process by imposing \textit{structured} ``brain-damage'' on the kernel tensor. We demonstrate that considerable speed-up of ConvNets can be obtained for a certain structure.


This structure is motivated by the observation that the majority of current implementations of generalized convolutions (including the most efficient one at the time of submission) \cite{Chellapilla06,Donahue14,Jia14,Chetlur14,Vedaldi14,nervana} compute generalized convolutions by reducing them to matrix multiplications (this reduction is also referred to as \textit{lowering}, \textit{unrolling}, or the \texttt{im2col} operation). While unstructured brain damage in a convolutional layer, i.e.\ shrinking some of the coefficients of the convolutional kernel tensor to zero, will make one of the factor matrices (\textit{the filter matrix}) sparse, it will not make the overall multiplication run faster. Our idea therefore is to group together the entries of the convolutional tensor in a certain fashion and to shrink such groups to zero in a coordinated way. By doing this, we can eliminate rows and columns from both factor matrices that are multiplied when convolution is reduced to matrix multiplication. Repeated elimination of rows and columns makes both factor matrices thinner (but still dense) and results in faster matrix multiplication. 

We demonstrate that conventional group sparsity regularizer \cite{Yuan06} embedded into stochastic gradient descent minimization is able to accomplish group-wise brain damage efficiently. The use of group sparsity thus allows us to optimize receptive fields in the convolutional network. Our approach therefore makes the case for the natural idea of using structured sparsity as a simple way to optimize connectivity in deep architectures.

In the experiments, we show that a carefully-designed group-wise brain damage procedure can sparsify existing neural networks considerably. In particular, speed-up factors exceed those obtained by recent tensor-factorization based methods. E.g., we show that group-wise brain damage can accelerate the bottleneck layers of AlexNet ('conv2' and 'conv3') by a factor of $8.5$x simultaneously, while incurring only modest (~$1\%$) loss of the prediction accuracy.

\section{Related work} 

As ConvNets are growing in size and are spreading towards real-time and large-scale computer vision systems, a lot of attention is attracted to the problem of speeding up convolutional layers. In parallel to the lowering-based approaches mentioned above, which reduce convolutions to matrix multiplications, several works investigate the use of fast Fourier transforms \cite{Mathieu13,Vasilache14}. Despite the theoretical appeal, the use of Fourier transforms has its own limitations (mostly related to memory usage) and most existing packages stick to the lowering approach, which at the moment of the submission is also used by the fastest implementation~\cite{nervana}.

Alternatively, several recent works investigate various kinds of tensor factorization in order to break generalized convolution into a sequence of smaller convolutions with fewer parameters \cite{Jaderberg14,Denton14,Lebedev15}. Using inexact low-rank factorizations within such approaches allows to obtain considerable speedup when low enough decomposition rank is used. Our approach is related to tensor-factorization approaches as we also seek to replace full convolution tensor with a tensor that has fewer parameters. Our approach however does not perform any sort of decomposition/factorization for the kernel tensor.
Another more distantly related approach is represented by a group of methods~\cite{Ba14,Hinton14,Romero14} that compress the initial large ConvNet into a smaller network with different architecture while trying to match the outputs of the two networks. 

Our approach is also related to methods that use structured sparsity \cite{Yuan06,Roth08,Jenatton11} to discover optimal architectures of certain machine learners, e.g.\ to discover the optimal structure of a graphical model \cite{Jalali11} or the optimal receptive fields in the two layered image classifier \cite{Jia12}. On the other hand, since our approach effectively learns receptive fields within a ConvNet, it can be related to other receptive field learning approaches, e.g.~\cite{Coates11,Malinowski13}.

The combination of sparsity and deep learning has been investigated within several unsupervised approaches such as sparse autoencoders~\cite{Bengio07,Boureau08} and sparse deep belief networks~\cite{Lee08}. We also note two reports that use some form of sparsification of deep feedforward networks and appeared in the recent months as we were developing our approach. Similarly to \cite{LeCun90}, the work \cite{Collins14} uses sparsification to reduce the number of parameters in the memory-bound scenario. Their goal is thus to save memory rather than to attain acceleration. In the report of \cite{Figurnov15}, the output of the convolution is computed at a sparsified set of locations with the gaps being filled by interpolation. This approach does not sparsify the convolutional kernel and is therefore different from the group-wise brain damage approach we suggest here.

Our work focuses on the task of speeding up convolutional layers (as they represent the speed bottleneck) and is therefore complimentary to approaches that focus on the reduction of size/memory footprint of fully-connected layers~\cite{Chen15,Gupta15,Novikov15,Sainath13,Xue13}.

\section{Group-Sparse Convolutions}

\begin{figure*}
\centering
\noindent
\includegraphics[width=16cm]{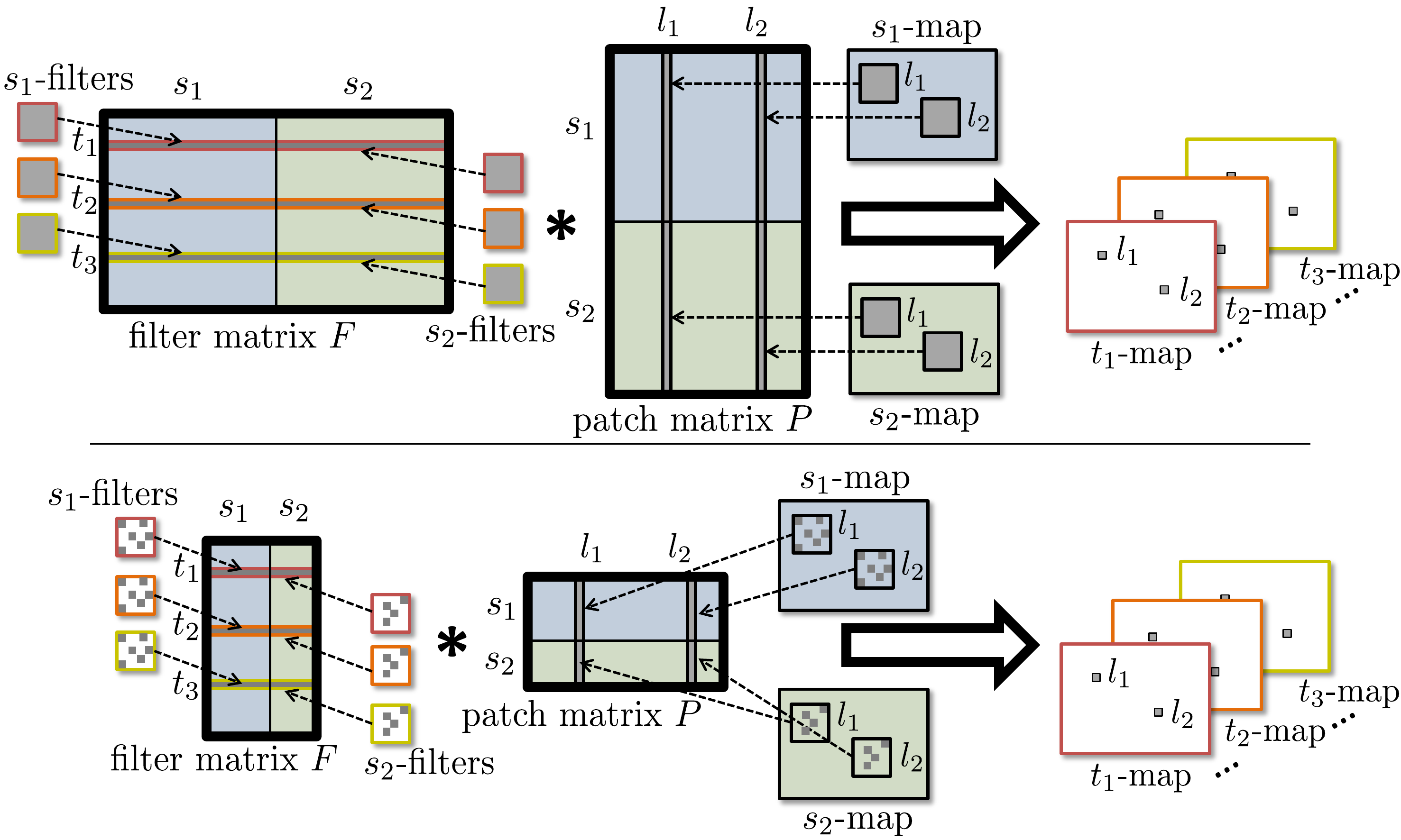}
\caption{Standard Generalized Convolution (top) vs.\ Generalized Convolution after Group-wise Brain Damage (bottom). In both cases, we show the diagram for two input maps ($S=2$, blue-green color coding). We highlight three output maps $t_1,t_2,t_3$ color-coded red-orange-yellow, and we also highlight two spatial locations $l_1$ and $l_2$. In both cases, the output map stack is obtained by reshaping the product of the filter matrix and the patch matrix. In the standard case, the filters and the patches sampled during the formation of the patch matrix are dense. After group-wise brain damage, both the filters and the patch sampling patterns are group-sparse (one sparsity pattern per input map), which results in much thinner filter and patch matrices and thus leads to much faster matrix multiplication/convolution.}
\label{fig:scheme}
\end{figure*}

Below, we discuss the reduction from generalized convolution to matrix multiplication \cite{Chellapilla06} and introduce the notation along the way. We then explain the group-sparse convolution idea.
Generalized convolution within a convolutional layer transforms an input stack of $S$ maps of size $W'{\times}H'$, which can be treated as a three-dimensional tensor (array) $U_{whs}$, into an output stack of $T$ maps of size $W''{\times}H''$ which form a three-dimensional tensor $V_{wht}$. The exact relation between $W',H'$ and $W'',H''$ depends on the padding and stride settings within the layer, and our approach can handle any padding/striding settings seamlessly. The transformation is defined by the following formula:

\newcommand{\halfd}{{\mbox{$\frac{d+1}{2}$}}}

\begin{align} \label{eq:convfull}
V(x,y,t) = \sum_{s=1}^S \sum_{\substack{i=1..d\\j=1..d}}& \; K(i,j,s,t)\,\cdot\\ \, & U(x{+}i{-}\halfd, y{+}j{-}\halfd, s) \nonumber
\end{align}

Here, $K$ is a four-dimensional \textit{kernel tensor} of size $d{\times}d{\times}S{\times}T$ with the first two dimensions corresponding to the spatial dimensions, the third dimension corresponding to input maps, the fourth dimension corresponding to output maps. The spatial width and height of the kernel are denoted as $d$ (for simplicity, we assume square shaped kernels and odd $d$).

The implementation of \eq{convfull} constitutes the speed bottleneck for ConvNets. In \cite{Chellapilla2006}, it was suggested to reduce the computation of all entries of $V$ to the multiplication of two large and dense matrices. The reduction allows to use highly optimized implementations of dense matrix multiplications (e.g.\ variants of BLAS~\cite{Blackford02} libraries) that have been developed over many years for all possible computing architectures.  The reduction proceeds as follows:

\begin{itemize}
\item The kernel tensor $K$ is reshaped into the \textit{filter matrix} $F$ of size $T \times d^2S$, where the $t$-th row corresponds to a sequence of $S$ 2D filters $K(:,:,s,t)$ reshaped in a row-wise fashion into row vectors.

\item The input map stack $V$ is reshaped into the \textit{patch matrix} $P$ of size $d^2S \times W''H''$, where the $l$-th column corresponds to a certain output location $l=(x,y)$ and is stacked from the $S$ patches extracted from $S$ input maps, all centered at this location and reshaped in a row-wise fashion into column vectors.

\item The filter matrix $F$ is multiplied by the patch matrix $P$ resulting in a matrix $\tilde V$ of size $T \times W''H''$ that contains all the elements of $V$ (each column corresponds to a certain location and contains the values of this location in the $T$ output maps). The multiplication implements \eq{convfull} exactly, as each row-by-column product within the multiplication corresponds to one instance of the computation \eq{convfull} for certain $(x,y,t)$. The output tensor (map stack) $V$ can be obtained from $\tilde V$ by reshaping.
\end{itemize}

The construction discussed above has proven to be highly successful and is used in the majority of modern ConvNets ``backends'', e.g.\ \cite{Chellapilla2006,Donahue14,Jia14,Chetlur14,Vedaldi14,nervana}. Our key idea is to train ConvNets with sparse convolutional kernels that are consistent with this construction.

Such consistency can be achieved if the sparsity patterns are aligned in a certain way. Formally, group-wise brain damage introduces a \textit{sparsity pattern} $Q_s$ for every input map $s \in 1\dots S$. The sparsity pattern is defined as a subset of the full spatial $d$-by-$d$ grid, i.e.\ $Q_s \subset \{1\dots d\} \otimes \{1\dots d\}$. The convolutional operation then becomes a slight modification of \eq{convfull}:

\begin{align}\label{eq:convsparse}
V(x,y,t) = \sum_{s=1}^S \sum_{\mathbf (i,j) \in Q_s}& \; K(i,j,s,t)\,\cdot\\ 
   & U(x{+}i{-}\halfd, y{+}j{-}\halfd, s) \nonumber
\end{align}

The reduction of \eq{convsparse} is an almost straightforward replication of the procedure \cite{Chellapilla2006}. The only modifications are (\fig{scheme}):
\begin{itemize}
\item When the filter matrix is assembled, each 2D filter $K(:,:,s,t)$ is reshaped into a row-vector of length $|Q_s|$ by including only non-zero elements. The filter matrix thus becomes of size $T \times \sum_{s=1}^S |Q_s|$.

\item When the patch matrix is assembled, each 2D patch at location $l=(x,y)$ in map $S$ is reshaped into a column vector of size $|Q_s|$ by sampling the input map $U(:,:,s)$ sparsely at locations $(x{+}i{-}\halfd, y{+}j{-}\halfd)$, where $(i,j) \in Q_s$. The patch matrix thus becomes of size $\sum_{s=1}^S |Q_s| \times W''H''$.
\end{itemize}

\begin{figure}
\centering
\setlength\figureheight{0.3\textwidth}
\setlength\figurewidth{0.4\textwidth}
%
%
%
%
\begin{tikzpicture}

\begin{axis}[
xlabel={density},
ylabel={relative time},
xmin=0, xmax=1,
ymin=0, ymax=1,
axis on top,
width=\figurewidth,
height=\figureheight
]
\addplot [black]
coordinates {
(2.77555756156289e-17,0)
(1,1)

};
\addplot [very thick, green!50.0!black, mark size=3,
  error bars/.cd,
  y dir=both,
  y explicit,
  error bar style={line width=1.5pt, xshift=0.0mm},
  error mark options = {
  line width=1.5pt, 
  mark size=3pt,
  }]
coordinates {
( 1.0 , 1.02943239545 ) +- ( 0 , 0.0252918561079 )
( 0.9 , 0.986141502735 ) +- ( 0 , 0.0494034498073 )
( 0.8 , 0.794324798113 ) +- ( 0 , 0.0146656213681 )
( 0.7 , 0.726689847377 ) +- ( 0 , 0.0149706055307 )
( 0.6 , 0.595360412601 ) +- ( 0 , 0.00741820617337 )
( 0.5 , 0.535082732741 ) +- ( 0 , 0.0172130598011 )
( 0.4 , 0.408741152181 ) +- ( 0 , 0.00775203803156 )
( 0.3 , 0.319203360589 ) +- ( 0 , 0.00388466878286 )
( 0.2 , 0.214212088193 ) +- ( 0 , 0.00449650421106 )
( 0.1 , 0.130030526775 ) +- ( 0 , 0.00809502100965 )
( 0.0 , 0.0262123708958 ) +- ( 0 , 0.0005540336466 )
};
\path [draw=black, fill opacity=0] (axis cs:13,1)--(axis cs:13,1);

\path [draw=black, fill opacity=0] (axis cs:1,13)--(axis cs:1,13);

\path [draw=black, fill opacity=0] (axis cs:13,0)--(axis cs:13,0);

\path [draw=black, fill opacity=0] (axis cs:2.77555756156289e-17,13)--(axis cs:2.77555756156289e-17,13);

\end{axis}

\end{tikzpicture}
\caption{Relative speed-up (with error bars over 100 runs) versus density level $\tau$, measured for forward propagation in the second convolutional layer of LeNet on CPU. Importantly, the observed speedup is almost linear in the sparsity level (diagonal).}
\label{fig:speedup}
\end{figure}
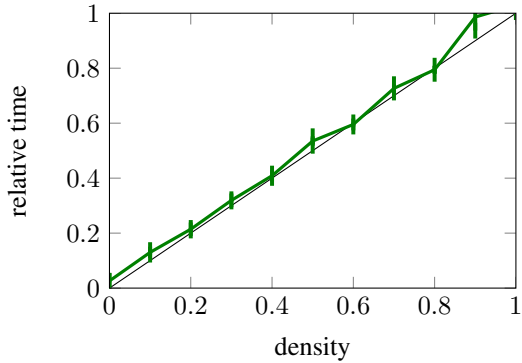

As a result of this modification, the multiplication of two dense matrices of sizes $T \times d^2S$ and $d^2S \times W''H''$ is replaced by the multiplication of two dense matrices of sizes $T \times \sum_{s=1}{S} |Q_s|$ and $\sum_{s=1}^S |Q_s| \times W''H''$, which results in the $d^2S/\sum_{s=1}^S |Q_s|$-times reduction in the number of scalar operations. In our experiments with the reference implementation of \cite{caffe} the wall-clock reduction in the convolution time between the original implementation and the group-sparse convolution was almost matching the ``theoretical'' speed-up factor (\fig{speedup}).

\section{Fast ConvNets with Group-sparse convolutions}

We consider two different scenarios that obtain fast ConvNets with group-sparse convolutions. First, we consider training such networks from scratch, and secondly we consider obtaining such networks by modification of pretrained architectures (i.e.\ performing ``brain damage'').

\subsection{Training from scratch}
\label{sec:scratch}

{\bf Predefined group-sparsity pattern.} The simplest solution that we consider is to choose the sparsity patterns $\Omega_S$ in advance in a data-independent manner, and enforce these patterns during the learning of the network. One particular case of this approach is simply reducing the spatial size of filters to a minimum, e.g.\ three-by-three, or even smaller rectangular pattern all the way to one-by-one (this is in line with a recent work of \cite{He15} where they consider $2\times2$ filters for some of their architectures). Note, that with our approach we are free to choose non-rectangular filters, and in the experiments we found this very useful.

One of the downsides of this approach is that when designing an architecture with multiple convolution layers, there are no clear design principles that can guide the choice of the filter shapes. In contrast, the methods discussed below can start with larger filters and then shrink their sizes towards optimally-shaped small filters. 

{\bf Training with group-sparsity regularizer.} Rather than fixing the group-sparsity pattern in advance, it is possible to find it as a part of learning process while the network is trained. A classical way to achieve this is through the use of group-sparsity regularization~\cite{Yuan06,Roth08,Jenatton11}. Thus, we consider a regularizer based on $l_{2,1}$-norm:
\begin{equation}
\Omega_{2,1}(K) = \lambda \sum_{i,j,s} \|\Gamma_{ijs}\| = \lambda \sum_{i,j,s} \sqrt{ \sum_{t=1}^T K(i,j,s,t)^2 }\,, \label{eq:l12}
\end{equation}
where the vector $\Gamma_{ijs}$ denotes the group of kernel tensor entries $K(i,j,s,:)$. The effect of the regularizer \eq{l12} is in shrinking some of such groups to zero in a coordinated fashion. When an entire group $\Gamma_{ijs}$ is set to zero, one can set the pixel $(i,j)$ in the sparsity pattern $\Omega_s$ to zero, thus increasing the group-sparsity.

For a convolutional layer that is being sparsified, the gradient of \eq{l12}, i.e.:
\begin{equation}\label{eq:grad}
\frac{ \partial \Omega_{2,1}(K)} {\partial K(i,j,s,t)}  = \lambda \frac{K(i,j,s,t)}{ \sqrt{ \sum_{z=1}^{T} K(i,j,s,z)^2 } }
\end{equation}
can simply be added to the gradient of the learning loss while performing stochastic gradient updates in the course of learning. The coefficient $\lambda$ in \eq{l12} and \eq{grad} controls the strength of the regularization w.r.t.\ the main learning loss.

Generally, using the regularizer \eq{l12} will result in a  group-sparsified kernel tensor with some of $\Gamma_{ijs}$ having only near-zero entries. Because of the stochastic nature of SGD and non-differentiability of $l_{12}$ norm near zero, the entries in these groups will not be exactly zero, and further postprocessing is needed to nullify the near zero groups and to set the sparsity patterns $\Omega_S$ accordingly.

\subsection{Sparsifying with Group-wise Brain Damage}

While it is possible to train ConvNets with group-sparse convolutions from scratch, the main focus of our paper is developing algorithms that can speed-up existing pretrained networks that often take excessive time for training. Towards this end, we have developed two approaches that can accelerate pretrained networks by inflicting group-wise brain damage in a way that the drop in the prediction accuracy is kept small. In both cases, we assume that we have  access to the training dataset $D$, the model was trained on.

{\bf Group-wise sparsification with fine-tuning.} Our first implementation is also based on the group-sparse regularizer \eq{l12}. We start with the input ConvNet and run the learning process on the dataset $D$ with the added regularizer \eq{l12}. After a certain amount of iterations, a predefined number of groups $\Gamma_{ijs}$ with the smallest $l2$-norm is set to zero. For a desired density level $\tau \in [0,1]$ and respective speedup $1/\tau$, we set $d^2S\,(1-\tau)$ groups to zero, making the respective $Q_S$ sparse. 

We have found two complications with this approach. Firstly, for a given density $\tau$ it was generally hard to set appropriate regularization strength $\lambda$ in advance without trying several values. Secondly, small $\tau$ (large speedup) the appropriate regularization strength $\lambda$ typically leads to an excessive regularization, as many groups end up being biased towards zero but not close to zero. Because of that, the prediction accuracy for such $\lambda$ experienced significant drop in the process of learning as compared to the input ConvNet.

Fortunately, one can recover from  most of this drop by the subsequent \textit{fine-tuning} of the network, that follows after the brain-damage process. For the fine-tuning, we fix the sparsity patterns $Q_S$ and restart learning without group-sparse regularization. We then train for an excessive number of epochs. As a result of such fine-tuning, the network adapts to the imposed sparsity patterns, while the prediction accuracy goes up and recovers most of the drop. 

{\bf Gradual group-wise sparsification.} To avoid the two complications discussed above we developed an alternative approach that essentially combines the brain-damage and the fine-tuning processes, and furthermore avoids most of the need for manual search for good meta-parameter values. The approach also often leads to considerably better results.

In this approach, we consider the \textit{truncated} $l_{12}$ regularizer:
\begin{equation} \label{eq:trunc}
\Omega^T_{2,1}(K) = \lambda \sum_{i,j,s} \min(\|\Gamma_{ijs}\|, \theta)
\end{equation}
The gradient of \eq{trunc} equals \eq{grad} when $\|\Gamma_{ijs}\| < \theta$ and is zero otherwise. Informally speaking, the value of $\theta$ controls which groups are considered ``promising'' and are being shrinked towards zero, and which groups are considered to be too far from zero and therefore stay unaffected by the regularizer \eq{trunc}.

To perform brain-damage, we then create a validation set on which we monitor the performance of the network. We choose the maximum drop $\delta$ of the prediction accuracy on the validation set that we are willing to tolerate. We then start with an input ConvNet and perform  learning with the regularizer \eq{trunc} while varying $\theta$. Specifically, after each epoch we monitor the performance of the network on a hold-out set and increase $\theta$ (intensifying brain damage) if the accuracy drop is less than $\delta$ and decrease $\theta$, thus relieving certain groups from the effect of the regularizer, if the drop is greater than $\delta$. 

To perform the actual sparsification, we also introduce an additional threshold $\epsilon{\ll} \delta$. In the process of learning, when the norm of a certain group falls below the threshold (i.e. $|\Gamma_{ijs}\| < \epsilon$) the group is greedily fixed to zero and eliminated from the tensor. The sparsity thus monotonically increases through the process, and we carry on training until the sparsification process stalls, i.e.\ the system keeps training with $\Gamma$ and performance drop oscilating, while no new groups have their legths fall under $\epsilon$ for a  number of epochs. In our experiments, all increments and decrements of $\theta$ was based on five-percent quantiles of the groups. I.e.\ every time $\theta$ is adjusted, we set $\theta$ to bring $5\%$ of groups $\Gamma_{ijs}$ in or out of the $\|\Gamma_{ijs}\|{<}\theta$  ``territory''. 

Overall, we found the whole procedure to be rather insensitive to the choices of $\lambda$ and $\epsilon$, and overall to be more practical and lead to higher group-sparsity and speed-ups than those attainable by the sparsification with fine-tuning approach. Most importantly, we could use same $\lambda$ and $\epsilon$, as well as same shared value of $\theta$ when sparsifiying multiple layers simultaneously.

\begin{figure}
\begin{minipage}{\linewidth}
    \centering
    \setlength\figureheight{0.7\textwidth}
    \setlength\figurewidth{1.0\textwidth}
%
%
%
%
\begin{tikzpicture}

\begin{axis}[
title={MNIST comparison},
xlabel={group density (\%)},
xmin=0, xmax=11,
ymin=0.9, ymax=1,
xtick={0, 2, 4, 6, 8, 10},
axis on top,
width=\figurewidth,
height=\figureheight,
legend entries={{$l_{2,1}$-regularizer},{$l_1$-regularizer},{fixed}},
legend style={at={(0.97,0.03)}, anchor=south east}
]
\addplot [thick, green!50.0!black]
coordinates {
(10.0392156862745,0.97969)
(9.05882352941176,0.97967)
(8,0.9788)
(7.01960784313725,0.97879)
(6.03921568627451,0.9788)
(5.05882352941177,0.97879)
(4,0.9441)
(3.01960784313726,0.94288)
(2.03921568627451,0.9291)
(1.05882352941177,0.75447)
(0.274509803921569,0.17171)

};
\addplot [thick, blue]
coordinates {
(10.0392156862745,0.97005)
(9.05882352941176,0.97016)
(8,0.95648)
(7.01960784313725,0.95647)
(6.03921568627451,0.95547)
(5.05882352941177,0.95566)
(4,0.74955)
(3.01960784313726,0.68211)
(2.03921568627451,0.50375)
(1.05882352941177,0.27279)
(0.274509803921569,0.1459)

};

\addplot [thick, black, only marks]
coordinates {
(8.0,0.9690675)
(4.0,0.92586)
};
\path [draw=black, fill opacity=0] (axis cs:13,1)--(axis cs:13,1);

\path [draw=black, fill opacity=0] (axis cs:0.12,13)--(axis cs:0.12,13);

\path [draw=black, fill opacity=0] (axis cs:13,0)--(axis cs:13,0);

\path [draw=black, fill opacity=0] (axis cs:3.46944695195361e-18,13)--(axis cs:3.46944695195361e-18,13);

\end{axis}

\end{tikzpicture}


    
    \caption{Accuracy vs. density level on MNIST dataset (LeNet architecture) for various ConvNets with group-sparse convolutions. We compare the results obtained by training with $l_{2,1}$ and $l_1$ regularizations followed by sparsifications, as well as training with predefined sparsity patterns $\Omega_S$ (black dots). Overall, training with $l_{2,1}$ regularizer obtains the best result that can be further improved by fine-tuning without regularization.}
    \label{fig:lenet}
\end{minipage}
\end{figure}
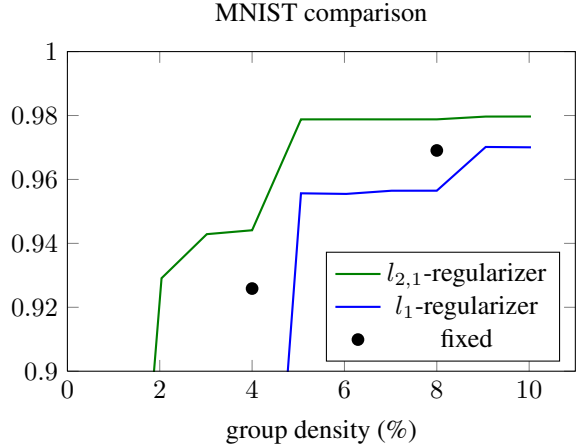

\begin{table*}[]
\centering
\begin{tabular}{|c|c|c|c|c|c|}
\hline
 \textbf{method} & \textbf{density} & \textbf{speed-up} & \textbf{accuracy drop} \\
  \hline
  \multicolumn{4}{|c|}{\bf Accelerating the second convolutional layer of AlexNet} \\
 \hline
 Denton et al. \cite{Denton14}: Tensor decomposition + Fine-tuning & & 2.7x & $\sim{}1\%$\\
 Lebedev et al. \cite{Lebedev15}: CP-decomposition + Fine-tuning & & 4.5x & $\sim{}1\%$\\
 Jaderberg et al. \cite{Jaderberg14}: Tensor decomposition + Fine-tuning & & 6.6x & $\sim{}1\%$\\
  \hline
  Training with fixed sparsity patterns & 0.12 & 8.33  & 0.82\% \\
  Training with fixed sparsity patterns & 0.2 & 5x & 0.16\% \\   
  \hline
 Group-wise sparsification + Fine-tuning & 0.1 & 10x & 1.13\%  \\
 Group-wise sparsification + Fine-tuning & 0.2 & 5x & 0.43\%  \\
 Group-wise sparsification + Fine-tuning & 0.3 & 3.33x & 0.11\%  \\
 Group-wise sparsification + Fine-tuning & 0.4 & 2.5x & -0.09\% \\
 \hline
Gradual group-wise sparsification & 0.11 & 9.0x & 0.28\% \\
Gradual group-wise sparsification & 0.05 & 20x & 1.07\% \\
 \hline
 \multicolumn{4}{|c|}{\bf Accelerating the second and the third convolutional layers of AlexNet} \\
 \hline
 Training with fixed sparsity patterns & 0.12 & 8.7x & 1.54\% \\
 Training with fixed sparsity patterns & 0.35 & 2.9x & 0.36\% \\  
Training with fixed sparsity patterns & 0.54 & 1.9x & -0.53\% \\ 
 \hline
Group-wise sparsification + Fine-tuning & 0.2 & 5x & 1.50\% \\
Group-wise sparsification + Fine-tuning & 0.3 & 3.33x & 1.17\% \\
Group-wise sparsification + Fine-tuning & 0.5 & 2x & 0.57\% \\
 \hline
Gradual group-wise sparsification & 0.12 & 8.5x & 1.04\% \\ 
 \hline
 \multicolumn{4}{|c|}{\bf Accelerating all five convolutional layers of AlexNet} \\
 \hline
Training with fixed sparsity patterns & 0.34 & 3.0x & 1.34\% \\
 \hline
Gradual group-wise sparsification & 0.31 & 3.2x & 1.43\% \\ 
 \hline
\end{tabular}
\caption{\textbf{Accelerating convolutional layers of the pretrained AlexNet architecture}: results of the two variants of our method for various sparsity levels alongside tensor-decomposition based methods (note: the results for \cite{Jaderberg14} are reproduced from \cite{Lebedev15}). }
\label{tab:alexnet}
\end{table*}

\section{Experiments}

{\bf Implementation details.} Our implementation is based on \texttt{Caffe} \cite{caffe} and modifies their original convolution, which is implemented as two subsequent layers (the \texttt{im2col}-layer that forms the patch matrix and the multiplication layer). To implement the group-sparse convolution we focused on the forward propagation step and CPU computation. Most of our methods can be extended for backprop step and for GPUs, however making such extensions efficient is non-trivial. For our purpose, we only needed to modify the \texttt{im2col}-layer, so that it can fill in the patch matrix while following certain sparsity patterns. 

\textbf{Datasets.} We perform the following experiments. Firstly, we consider a small-scale setting, and compare training ConvNets with group-wise brain damage from scratch with baselines. We use MNIST dataset~\cite{LeCun98} for these small-scale experiments. We then consider a large-scale problem, namely ImageNet (ILSVRC) image classification and the task of accelerating of a pretrained architecture, namely the Caffe version of AlexNet~\cite{Krizhevsky12}.
We also give preliminary results for one of the VGGNet networks~\cite{Simonyan14}.

\subsection{ MNIST experiments}

We trained the LeNet architecture on the MNIST dataset from random initialization while adding the group-sparce regularization (section~\ref{sec:scratch}) while varying the regularization strength $\lambda$ and picking the optimal one for each sparsity level. The sparsification affects both convolutional layers of LeNet, and the same density level $\tau$ is enforced in both layers. We also consider a number of baselines:
\begin{itemize}
\item A simple baseline that trains the network without regularization and then simply eliminates (set to zero) a certain number of groups $\Gamma_{ijs}$ with the smallest l2-norms. The performance of this baseline was clearly below all other methods and it is not reported.
\item Picking sparsity patterns $Q_S$ in advance. We consider filters with only one central non-zero entry and filters with two adjacent central non-zero elements. These options correspond to the density of 4\% and 8\% respectively. The former is essentially equivalent to a non-convolutional network.  
\item We also consider a simpler non-group-wise sparsification by training with l1-norm regularizer (with varying $\lambda$) but then nullifying groups $|\Gamma_{ijs}$ based on their norms.
\end{itemize}
The results of the proposed method and the baselines are shown in \fig{lenet}. The rightmost plot shows the comparison of the $l_1$-envelope, $l_{2,1}$-envelope, and the performance of the group-wise brain damage applied to the network trained without sparsity-inducing regularizer. The use of group-sparsity regularization boosts the performance of group-wise brain damage very considerably. Twenty-fold acceleration of convolutional layers can be obtained while keeping the error low ($2.1\%$, reduced to $1.71\%$ after fine-tuning). Using $l_1$-regularizer followed by optimal brain damage works worse than $l_{2,1}$-regularizer. Pre-fixing sparsity patterns achieves good results, which are still worse than training with grou-sparsity regularizer. Note also that all methods except the baseline with the pre-fixed patterns can be improved via fine-tuning.


\subsection{ILSVRC experiments}

\begin{figure*}
\centering
\begin{subfigure}[b]{0.32\textwidth}
\centering
\includegraphics[width=\linewidth]{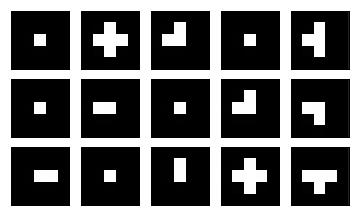}
\caption{sparsity $1-\tau=0.9$}
\end{subfigure}
\begin{subfigure}[b]{0.32\textwidth}
\centering
\includegraphics[width=\linewidth]{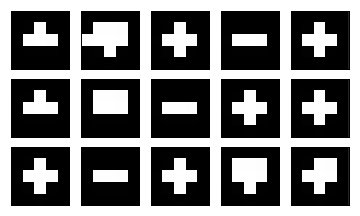}
\caption{sparsity $1-\tau=0.8$}
\end{subfigure}
\begin{subfigure}[b]{0.32\textwidth}
\centering
\includegraphics[width=\linewidth]{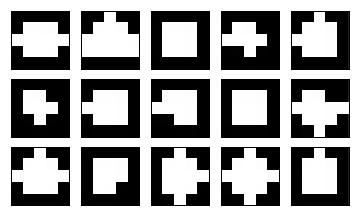}
\caption{sparsity $1-\tau=0.6$}
\end{subfigure}
\caption{The sparsity patterns obtained by group-wise brain damage on the second convolutional layer of AlexNet for different sparsity levels. Nonzero weights are shown in white. In general, group-wise brain damage shrinks the receptive fields towards the center and tends to make them circular.}
\label{fig:patterns}
\end{figure*}


We first consider the AlexNet (Caffe reimplementation) architecture that has five convolutional layers. We consider the following subtasks: (i) accelerating the second convolutional layer (which is the slowest of all layers), (ii) accelerating the second and the third layers (which are the two slowest layers), (iii) accelerating all five convolutional layers (which together take the vast majority of the forward-propagation time). When reporting the final density in subtasks (ii) and (iii), we weigh the densities in different layers by the forward propagation times.

We focused on accelerating the existing network from Caffe zoo (\tab{alexnet}). It is not clear how initializing network with pretrained weights as opposed to a random initialization affects the final accuracy, but it allowed to shorten training time in many cases, which is critical in case of large networks. We evaluate the variant of our method that trims the network according to some predefined sparsity pattern, and then learns the network while keeping the same fixed pattern. Namely we consider the following symmetric centered patterns: vertical or horizontal block 1$\times$3, the 3$\times$3 cross pattern, 3$\times$3 square or diamond shape inside 5$\times$5 filter. 



For the first two subtasks, we evaluated the variant of our method with sparsity-inducing regularizer for various sparsity levels. For several desired density levels $\tau$ we searched for optimal $\lambda$ through large range with ten-fold increments. For each $\tau$ we pick $\lambda$ that results in the minimal accuracy drop after sparsification before fine-tuning. After picking the optimal $\lambda$, we perform fine-tuning. \fig{patterns} demonstrate sparsity patterns $\Omega_S$ obtained for different sparsity levels.

Finally, for all three subtasks we evaluated the most advanced of our methods, namely gradual group-wise sparsification. We set the parameters $\lambda$ and $\epsilon$ to 0.01 and 0.1 respectively. We split the test set of ILSVRC randomly into two halves and use one of the halves solely to estimate the drop of the classification accuracy in the dynamical adjustment of $\theta$. We then report the performance drop on the other half of the test set. We set the acceptable performance drop to be $1\%$ of top-1 accuracy. 

\begin{figure*}
    \centering
    \setlength\figureheight{0.4\textwidth}
    \setlength\figurewidth{0.35\textwidth}
    \input{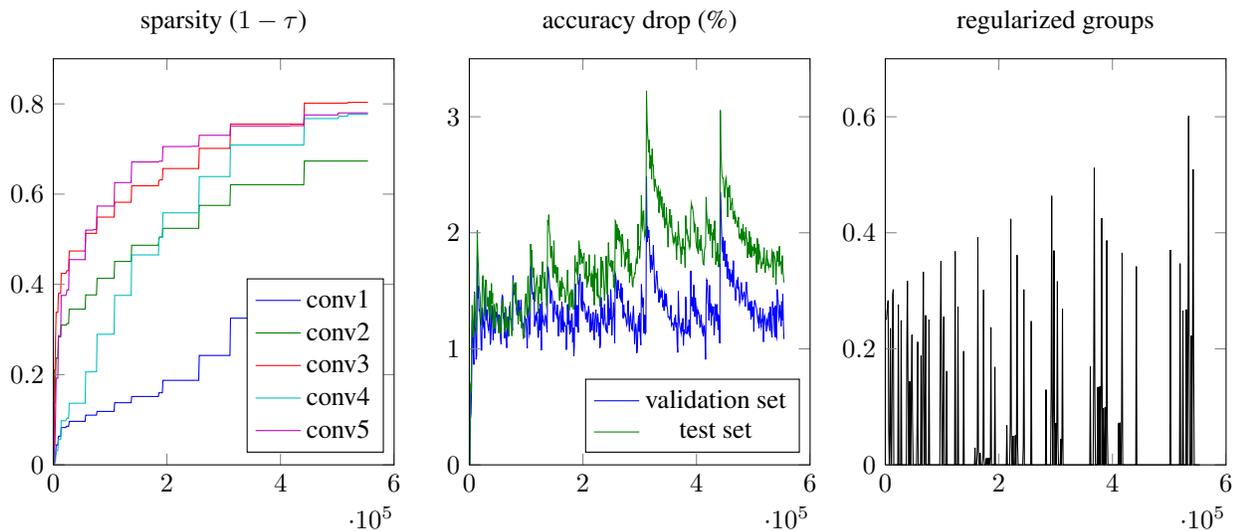} 
    \caption{The process of sparsification of all five layers in AlexNet. The left plot shows the monotonic growth of the sparsity levels of the five convolutional layers as the iterations progress. The middle plot shows the relative prediction accuracy drop for the current system for the validation part and for the hold-out test set. Finally, the right part visualizes the process of the adjustment of $\theta$ threshold in the truncated $l_{2,1}$ regularization. This plot shows the percentile of groups $\Gamma_{ijs}$ with the $l_2$-norm less than $\theta$. $\theta$ is increased or decreased dependent on whether the performance drop on the validation set is greater or smaller than $1.2\%$. } 
\label{fig:alexnet}
\end{figure*}

As shown in~\tab{alexnet}, the results of gradual sparsification outperform the tensor factorization methods as well as sparsification with fine-tuning considerably, achieving higher group-sparsification/speed-up for similar prediction accuracy drop. Notably, the proposed approach is more successful in speeding-up AlexNet than a number of approaches based on tensor decomposition. \fig{alexnet} further visualizes the process of the simultaneous gradual brain damage inflicted on all five layers of AlexNet.

{\bf ``External'' computer vision task.} Convolutional layers of large networks pretrained on large annotated training sets such as ILSVRC can be used as universal spatially localized features in a variety of ways \cite{Liu15,Azizpour15}, which is particularly valuable for problems with considerably smaller training sets. Recently, \cite{Babenko15} showed that descriptors obtained by sum-pooling of the features that emerge in the last convolutional layer of a pretrained network can be used as state-of-the-art holistic descriptors for image retrieval. We followed their approach (that includes PCA whitening and normalization as postprocessing) to assess the effect of group-sparsification on an external task. Comparing AlexNet as a base model, and the network with the simultaneous group-sparisfication of all convolutional layers from \tab{alexnet} with $3.2x$ speedup, we have found a negligible drop in performance for the INRIA holidays dataset~\cite{Jegou08} from 0.783 mAP to 0.780 mAP, and a reasonably small drop for the Oxford Building dataset~\cite{Philbin07} from 0.45 to 0.41.

{\bf Preliminary VGGNet results.} We have also applied the gradual group-wise sparsification to the slowest convolutional layer of VGGNet (the deeper 19 layer version of \cite{Simonyan14}, starting from its Caffe Zoo version. The sparsification obtained the density $\tau=0.13$ with only $0.2\%$ top-1 accuracy drop. Interestingly, unlike the experiments with AlexNet where we rarely observed empty sparsity patterns $\Omega_S$ (``dead feature maps''), in this example such all-zero patterns were present (29 out of 64), suggesting that this manually designed architecture contains excessive number of feature maps in this layer. This result also suggest that our approach is suitable even for networks with very small initial filter sizes in convolutional layers ($3\times{}3$ for VGGNet).

\section{Discussion}

We have presented an approach to speeding up ConvNets that uses the group-wise brain damage process that sparsifies convolution operations. The approach takes into account the way generalized convolutions are reduced to matrix multiplications, and prune the entries of the convolution kernel in a groupwise fashion. The exact sparsity patterns can be learned from data using group-sparsity regularization. When applied after learning with such regularization and followed by fine-tuning, group-wise brain damage obtains state-of-the-art performance for speeding up ConvNets.

Aside from the practical value, the proposed approach also makes the case for the use of sparse learning for automated discovery of optimal network architectures, which is arguably one of the main unsolved problems in deep learning. In our case, group-sparse regularizer allows the model to discover optimal receptive fields (\fig{patterns}). It is interesting to see that the optimization process decided to shrink the receptive fields towards the center compared to the full version (which is consistent with the findings in \cite{Simonyan14,He15}). Perhaps, even more interesting is to see that in general, the learning process decided to make the receptive fields roughly circular. Also, the process treated AlexNet and VGGNet differently, eliminating entire feature maps by assigning their sparsity patterns $\Omega_S$ to zero maps in the latter case. Note that such elimination brings additional speedup (since the entire map needs not be computed in the previous layer). Such elimination can be explicitly encouraged within our approach using hierarchical group-sparsity regularizers~\cite{Bach09,Jenatton11}.

\bibliographystyle{ieee}
\bibliography{references}

\end{document}